%% file: main.tex
\documentclass[11pt]{article}

\usepackage[preprint]{acl}

\usepackage{times}
\usepackage{latexsym}
\usepackage{xspace}
\usepackage{hyperref}

\usepackage[T1]{fontenc}

\usepackage[utf8]{inputenc}

\usepackage{microtype}

\usepackage{inconsolata}

\usepackage{graphicx}
\usepackage{booktabs} 
\usepackage{amsmath}
\usepackage{amssymb}
\usepackage[most]{tcolorbox}
\tcbuselibrary{breakable}

\usepackage[table]{xcolor}
\usepackage{multirow}
\usepackage{bm}

\title{\textit{\large Predict, Reuse, and Repair:}\\ Accelerating Dynamic Sparse Attention for Long-Context LLM Decoding}

\newcommand{\systemname}{\textsf{PRR}\xspace}

\newcommand{\squishlist}{
    \begin{list}{$\bullet$}
        { \setlength{\itemsep}{0pt}      \setlength{\parsep}{0pt}
            \setlength{\topsep}{0.5pt}       \setlength{\partopsep}{0pt}
            \setlength{\listparindent}{-2pt}
            \setlength{\itemindent}{-5pt}
            \setlength{\leftmargin}{1.1em} \setlength{\labelwidth}{0em}
            \setlength{\labelsep}{0.2em} } }
    
\newcommand{\squishend}{
\end{list}  }

\newcounter{squishnum} 

\newcommand{\squishenum}{
    \begin{list}{\arabic{squishnum}.} 
        { \usecounter{squishnum}      
          \setlength{\itemsep}{0pt}
          \setlength{\parsep}{0pt}
          \setlength{\topsep}{0.5pt}
          \setlength{\partopsep}{0pt}
          \setlength{\listparindent}{-2pt}
          \setlength{\itemindent}{-5pt}
          \setlength{\leftmargin}{1.1em} 
          \setlength{\labelwidth}{0em}
          \setlength{\labelsep}{0.2em} } }

\newcommand{\squishenumend}{
\end{list}  }

\newtcolorbox{observationbox}[1]{
  enhanced,
  colback=gray!7,                       
  colframe=black!80,                    
  boxrule=0pt,                          
  leftrule=3pt,                         
  sharp corners,
  boxsep=4pt,
  left=8pt,
  right=8pt,
  top=6pt,
  bottom=6pt,
  fontupper=\small,                     
  title={\textbf{#1}},                  
  coltitle=black,
  attach title to upper={\quad}         
}

\author{
  \textbf{Tianyu Wang\textsuperscript{1}},
  \textbf{Gourav Rattihalli\textsuperscript{2}},
  \textbf{Aditya Dhakal\textsuperscript{2}},
  \textbf{Junbo Li\textsuperscript{1}},
\\
  \textbf{Zhiwei Ren\textsuperscript{1}},
  \textbf{Dejan Milojicic\textsuperscript{2}},
  \textbf{Longfei Shangguan\textsuperscript{1}}
\\
\\
  \textsuperscript{1}University of Pittsburgh, Pittsburgh, PA, USA \quad
  \textsuperscript{2}HPE Labs, Milpitas, CA, USA
}

\begin{document}
\maketitle
\begin{abstract}
Dynamic sparse attention (DSA) accelerates long-context LLM decoding by attending to only the top-$K$ KV blocks relevant to each query, but it introduces a serialized selection-to-attention dependency that emerges as a new latency bottleneck. We present \systemname, a speculate-reuse-repair runtime that exploits temporal locality in DSA selections to predict likely blocks, speculate the attention over them while selection is in flight, and incrementally repair missed blocks once the true selected set is known. \systemname uses a lightweight EMA-based predictor, a profiling-guided speculation budget that keeps speculative work off the critical path, and a FlashAttention-based repair kernel that folds missed blocks into the partial attention state using online-softmax statistics. Across long-context benchmarks and representative DSA methods, \systemname reduces per-token decoding latency by up to 40\% while preserving downstream task accuracy. 
Github:
\url{https://github.com/Tianyu9748/Incremental_FlashAttention}
\end{abstract}

\input{tex/introduction}
\input{tex/motivation}
\input{tex/design}
\input{tex/evaluation}
\input{tex/related_work}

\section*{Limitations}
\systemname leaves rooms for improvement.
Firstly, we only evaluate over training-free dynamic sparse attention mechanisms (i.e., Quest and InfLLM-v2).
We don't include any trainable mechanisms like NSA in this paper, which requires huge computing resources to train from scratch.
Secondly, we only optimize the kernel performance for a specific GPU architecture (i.e., NVIDIA Hopper) and under half precision.
Future work can extend our kernel optimization to other architectures (e.g., Blackwell) and other low-bit precisions (e.g., fp8).
Thirdly, we mainly present the results of GLM-4-9B in the main content due to the page limit.
We perform the same experiments for other LLMs and present the results in the Appendix~\ref{s:appendix}, where the same trend is observed on other LLMs as GLM-4-9B.
Lastly, we only evaluate \systemname of batch size 1. 
Future work can explore how to coordinate stages (e.g., block selection, speculation) across requests within the same batch to improve decoding throughput.
\bibliography{custom}

\appendix

\section{Appendix}
\label{s:appendix}

\subsection{Extend Quest to GQA}
The original Quest formulation targets Multi-Head Attention, where each query head has a dedicated KV head and independently selects its top-$K$ pages. Under Grouped-Query Attention (GQA), $G$ query heads share one KV head, making per-query-head selection both wasteful---selected pages are loaded once per KV head---and ill-defined when heads in a group disagree. We extend Quest to GQA by operating at KV-head granularity: page metadata (channel-wise min/max keys $m_p, M_p$) is stored per KV head as before, and for each KV head we form a representative query $\bar{Q} = \frac{1}{G}\sum_{g=1}^{G} Q_g$. Criticality estimation then proceeds as $s_p = \sum_i \max(\bar{Q}_i m_{p,i},\, \bar{Q}_i M_{p,i})$, yielding one score per (KV-head, page) from which top-$K$ pages are selected and shared across the group during sparse attention. 

\subsection{Setup for Quest and InfLLM-v2}
We use the same setting (e.g., block size, token budget) as Quest~\cite{quest} and InfLLM-v2~\cite{infllm-v2} for experiments in Section~\ref{s:eval}.

\subsection{Temporal Locality across LLMs and Benchmarks}
\label{ss:appdix_temporal_locality}

We measure how many blocks are repeatedly selected in consecutive decoding steps on five representative long-context benchmarks, including LongBench~\cite{longbench}, InfiniteBench~\cite{infinitebench}, AIME~\cite{hf_aime_2024,balunovic_srimatharena_2025}, MATH500~\cite{hf_math500}, and RULER~\cite{ruler} for GLM-Z1-9B~\cite{glm-z1}, Deepseek-R1-8B~\cite{deepseek-r1}, Llama3-8B-1M~\cite{llama3-8b-1m}, Qwen3-14B~\cite{qwen3}, and Qwen3-32B~\cite{qwen3}.

Table~\ref{tab:overall_overlap_rate} reports the profiled overlap rates. Across all LLMs and benchmarks, the overlap stays below 0.7. 
This suggests that naively reusing the previous step's block selection offers little room for speculative speedup, due to frequently triggered fallback attention recomputation.

\begin{table}[]
\centering
\footnotesize
\setlength{\tabcolsep}{0.6pt} 
\caption{Average overlap rate of blocks in consecutive block selections of LLMs under various benchmarks.}
\label{tab:overall_overlap_rate}
\begin{tabular}{lcccccc}
\toprule
\textbf{Model} & \textbf{\shortstack{Long\\Bench}} & \textbf{\shortstack{Infinite\\Bench}} & \textbf{RULER} & \textbf{AIME} & \textbf{\shortstack{MATH\\500}} & \cellcolor{gray!20}\textbf{Avg.} \\
\midrule
\multicolumn{7}{c}{\textbf{Quest}} \\
\midrule
GLM-4       & 0.669 & 0.694 & 0.658 & 0.674 & 0.664 & \cellcolor{gray!20}\textbf{0.672} \\ \midrule
GLM-Z1      & 0.689 & 0.693 & 0.700 & 0.695 & 0.707 & \cellcolor{gray!20}\textbf{0.697} \\ \midrule
DeepSeek-R1 & 0.629 & 0.639 & 0.638 & 0.633 & 0.639 & \cellcolor{gray!20}\textbf{0.636} \\ \midrule
Llama3      & 0.666 & 0.649 & 0.665 & 0.671 & 0.689 & \cellcolor{gray!20}\textbf{0.668} \\ \midrule
Qwen3-14B   & 0.640 & 0.655 & 0.662 & 0.652 & 0.657 & \cellcolor{gray!20}\textbf{0.653} \\ \midrule
Qwen3-32B   & 0.637 & 0.657 & 0.640 & 0.655 & 0.633 & \cellcolor{gray!20}\textbf{0.644} \\ 
\midrule
\midrule \addlinespace 
\multicolumn{7}{c}{\textbf{InfLLM-v2}} \\
\midrule
GLM-4       & 0.689 & 0.701 & 0.679 & 0.718 & 0.702 & \cellcolor{gray!20}\textbf{0.698} \\ \midrule
GLM-Z1      & 0.692 & 0.708 & 0.701 & 0.698 & 0.694 & \cellcolor{gray!20}\textbf{0.698} \\ \midrule
DeepSeek-R1 & 0.669 & 0.657 & 0.681 & 0.670 & 0.677 & \cellcolor{gray!20}\textbf{0.671} \\ \midrule
Llama3      & 0.679 & 0.639 & 0.672 & 0.695 & 0.699 & \cellcolor{gray!20}\textbf{0.677} \\ \midrule
Qwen3-14B   & 0.651 & 0.656 & 0.675 & 0.668 & 0.682 & \cellcolor{gray!20}\textbf{0.666} \\ \midrule
Qwen3-32B   & 0.646 & 0.642 & 0.663 & 0.644 & 0.680 & \cellcolor{gray!20}\textbf{0.655} \\ 
\bottomrule
\end{tabular}
\end{table}

\subsection{Low Utilization of DSA}
\label{ss:appdix_low_util}

We profile the utilization of streaming multiprocessors (SMs), L2 bandwidth, and DRAM bandwidth for GLM-4-9B at larger batch sizes (i.e., 2, 4, 8, 16) under Quest and InfLLM-v2. 
The total context length is fixed at 128K, so batch size 16 corresponds to a per-prompt length of 8K tokens --- well above the token budgets used by both Quest and InfLLM-v2. 
As shown in Table~\ref{tab:appdix_gpu_util}, GPU resources remain underutilized even at these larger batch sizes, leaving ample headroom for speculative attention to execute in parallel with block selection.

\begin{table}[]
\centering
\footnotesize
\setlength{\tabcolsep}{3.2pt}
\caption{Utilization (\%) of SM, L2 bandwidth, and DRAM bandwidth under varying batch sizes for GLM-4-9B under Quest and InfLLM-v2.}
\label{tab:appdix_gpu_util}
\begin{tabular}{l l cccc}
\toprule
\textbf{DSA} & \textbf{Metric} & \textbf{BS=2} & \textbf{BS=4} & \textbf{BS=8} & \textbf{BS=16} \\
\midrule
\multirow{3}{*}{\textbf{Quest}}
& SM      & 6.73  & 7.22  & 7.90  & 8.66  \\
& L2 BW   & 43.46 & 44.33 & 44.32 & 45.57 \\
& DRAM BW & 36.55 & 37.26 & 37.24 & 38.02 \\
\midrule
\multirow{3}{*}{\textbf{InfLLM-v2}}
& SM      & 6.55  & 7.31  & 7.93  & 9.27  \\
& L2 BW   & 42.53 & 43.90 & 45.84 & 47.36 \\
& DRAM BW & 35.73 & 36.71 & 38.40 & 39.50 \\
\bottomrule
\end{tabular}
\end{table}

\subsection{Search and Prediction Cost}
\label{ss:design_search_prediction_cost}
Each grid evaluation runs the predictor over the prefill score matrix at a cost of $O(T_p \cdot N_p \log K)$ arithmetic operations.
Crucially, the search is scheduled to overlap with the prefill phase: by the time the last prefill token is processed and $IS$ is complete, most grid evaluations have already executed concurrently on the CPU.
The critical-path cost of calibration is therefore the tail of the grid that cannot be overlapped.
End-to-end, the full search---covering both overlapped and non-overlapped portions---takes approximately 2.4\,ms and 11.2\,ms for 16K- and 64K-token prefills, respectively, on GLM-4-9B. 
The non-overlapped portion (i.e., the fit for the final layer) accounts for only 0.06\,ms and 0.28\,ms for the two prefills.
The prediction process takes only 0.01\,ms.

\subsection{Profiling Overheads}
\label{ss:appdix_profiling_overheads}
Profiling the execution time of block selection and speculative attention computation across context lengths (4K, 8K, 16K, 32K, and 64K) takes under 5 minutes on a single H100 GPU for GLM-4-9B. For context lengths outside this grid, we estimate execution time via polynomial regression, a standard technique for continuous resource modeling in LLM serving~\cite{wang2025using,stojkovic2025dynamollm,hu2025hedrarag}.

\subsection{EMA Hit Rate}
\label{ss:appdix_ema_hit_rate}
Table~\ref{tab:overall_ema_hit_rate} reports the overlap rate between EMA's predicted block set and the true top-$K$ set of LLMs across benchmarks. \systemname's EMA predictor consistently achieves a high overlap rate (>97\%).

\begin{table}[]
\centering
\footnotesize
\setlength{\tabcolsep}{0.6pt} 
\caption{The overlap rate(\%) between EMA's prediction and the true top-$K$ set of LLMs across benchmarks.}
\label{tab:overall_ema_hit_rate}
\begin{tabular}{lcccccc}
\toprule
\textbf{Model} & \textbf{\shortstack{Long\\Bench}} & \textbf{\shortstack{Infinite\\Bench}} & \textbf{RULER} & \textbf{AIME} & \textbf{\shortstack{MATH\\500}} & \cellcolor{gray!20}\textbf{Avg.} \\
\midrule
\multicolumn{7}{c}{\textbf{Quest}} \\
\midrule
GLM-4       & 98.65 & 96.86 & 98.36 & 98.15 & 98.25 & \cellcolor{gray!20}\textbf{98.05} \\ \midrule
GLM-Z1      & 98.55 & 97.33 & 97.95 & 97.93 & 98.09 & \cellcolor{gray!20}\textbf{97.97} \\ \midrule
DeepSeek-R1 & 98.56 & 97.60 & 96.99 & 97.71 & 97.88 & \cellcolor{gray!20}\textbf{97.75} \\ \midrule
Llama3      & 98.63 & 97.62 & 97.49 & 98.01 & 98.28 & \cellcolor{gray!20}\textbf{98.01} \\ \midrule
Qwen3-14B   & 97.49 & 98.91 & 98.17 & 97.93 & 98.10 & \cellcolor{gray!20}\textbf{98.12} \\ \midrule
Qwen3-32B   & 98.52 & 98.56 & 98.12 & 97.94 & 98.06 & \cellcolor{gray!20}\textbf{98.24} \\
\midrule
\midrule \addlinespace
\multicolumn{7}{c}{\textbf{InfLLM-v2}} \\
\midrule
GLM-4       & 97.86 & 97.14 & 96.46 & 98.15 & 98.01 & \cellcolor{gray!20}\textbf{97.52} \\ \midrule
GLM-Z1      & 99.12 & 98.89 & 98.56 & 98.42 & 98.49 & \cellcolor{gray!20}\textbf{98.70} \\ \midrule
DeepSeek-R1 & 99.03 & 99.01 & 98.27 & 99.40 & 98.57 & \cellcolor{gray!20}\textbf{98.86} \\ \midrule
Llama3      & 97.67 & 97.50 & 97.85 & 98.75 & 98.95 & \cellcolor{gray!20}\textbf{98.14} \\ \midrule
Qwen3-14B   & 98.37 & 98.63 & 98.66 & 98.56 & 98.62 & \cellcolor{gray!20}\textbf{98.57} \\ \midrule
Qwen3-32B   & 98.87 & 98.80 & 98.93 & 98.82 & 98.87 & \cellcolor{gray!20}\textbf{98.86} \\
\bottomrule
\end{tabular}
\end{table}

\subsection{Ablation Study}
\label{ss:appdix_ablation}

Table~\ref{tab:ablation_full} decomposes \systemname's end-to-end speedup into three cumulative stages (S1, S2, S3), evaluated across all six LLMs and five benchmarks under both Quest and InfLLM-v2, where S3 represents the standard \systemname configuration.

We make three observations.
First, naively enabling speculative attention through Stage-1 yields negligible improvement over serial DSA execution ($1.00$--$1.04\times$ across all configurations). Although the block selections reused from the previous decoding step overlap with the current true top-$K$ set by roughly 70\% on average, Stage-1 requires \emph{exact} coverage to be useful: a single missed block triggers a full attention recomputation over the true top-$K$ selection. Since complete overlap is rarely achieved in practice, the fallback recomputation negates the latency savings from speculation.
Second, Stage-2 introduces our customized sparse attention kernel, raising the average speedup to roughly $1.20\times$ under Quest and $1.20$--$1.32\times$ under InfLLM-v2. Beyond simply accelerating attention over non-contiguous blocks, the kernel enables incremental attention repair, so missed blocks can be patched in directly rather than triggering a costly full recomputation. The gain is larger under InfLLM-v2 because its sparse-attention stage carries more latency, leaving more headroom for the repair path to exploit.
Third, Stage-3 additionally activates the EMA-based predictor, which attains a 98\% overlap rate between the predicted and true top-$K$ sets (Table~\ref{tab:overall_ema_hit_rate}). This near-exact prediction sharply reduces the fraction of blocks that fall back to repair, lifting the average speedup to $1.28$--$1.42\times$ under Quest and $1.32$--$1.56\times$ under InfLLM-v2. The consistency of this stage-wise pattern across all six LLMs and five benchmarks indicates that \systemname's design generalizes rather than being tuned to any single configuration.

\begin{table*}[]
\centering
\footnotesize
\setlength{\tabcolsep}{2.5pt}
\caption{Ablation of decoding speedup across stages, models, and benchmarks. S1, S2, and S3 (\systemname) are reported relative to serial DSA execution ($1.00\times$). Higher is better.}
\label{tab:ablation_full}
\begin{tabular}{l ccc ccc ccc ccc ccc | ccc}
\toprule
& \multicolumn{3}{c}{\textbf{LongBench}}
& \multicolumn{3}{c}{\textbf{InfiniteBench}}
& \multicolumn{3}{c}{\textbf{RULER}}
& \multicolumn{3}{c}{\textbf{AIME}}
& \multicolumn{3}{c}{\textbf{MATH500}}
& \multicolumn{3}{c}{\cellcolor{gray!20}\textbf{Avg.}} \\
\cmidrule(lr){2-4}\cmidrule(lr){5-7}\cmidrule(lr){8-10}\cmidrule(lr){11-13}\cmidrule(lr){14-16}\cmidrule(lr){17-19}
\textbf{Model} & S1 & S2 & S3 & S1 & S2 & S3 & S1 & S2 & S3 & S1 & S2 & S3 & S1 & S2 & S3 & \cellcolor{gray!20}S1 & \cellcolor{gray!20}S2 & \cellcolor{gray!20}S3 \\
\midrule
\multicolumn{19}{c}{\textbf{Quest}} \\
\midrule
GLM-4       & 1.00 & 1.27 & 1.50 & 1.01 & 1.21 & 1.40 & 1.00 & 1.24 & 1.45 & 1.00 & 1.11 & 1.30 & 1.00 & 1.16 & 1.42 & \cellcolor{gray!20}1.00 & \cellcolor{gray!20}1.20 & \cellcolor{gray!20}\textbf{1.41} \\
GLM-Z1      & 1.00 & 1.23 & 1.48 & 1.01 & 1.20 & 1.39 & 1.00 & 1.22 & 1.43 & 1.00 & 1.15 & 1.33 & 1.00 & 1.20 & 1.45 & \cellcolor{gray!20}1.00 & \cellcolor{gray!20}1.20 & \cellcolor{gray!20}\textbf{1.42} \\
DeepSeek-R1 & 1.00 & 1.19 & 1.39 & 1.03 & 1.21 & 1.42 & 1.00 & 1.25 & 1.47 & 1.00 & 1.13 & 1.35 & 1.00 & 1.13 & 1.39 & \cellcolor{gray!20}1.01 & \cellcolor{gray!20}1.18 & \cellcolor{gray!20}\textbf{1.40} \\
Llama3      & 1.00 & 1.17 & 1.36 & 1.01 & 1.20 & 1.40 & 1.00 & 1.20 & 1.41 & 1.00 & 1.15 & 1.39 & 1.00 & 1.11 & 1.34 & \cellcolor{gray!20}1.00 & \cellcolor{gray!20}1.17 & \cellcolor{gray!20}\textbf{1.38} \\
Qwen3-14B   & 1.00 & 1.23 & 1.44 & 1.00 & 1.09 & 1.27 & 1.00 & 1.19 & 1.41 & 1.00 & 1.21 & 1.44 & 1.00 & 1.21 & 1.45 & \cellcolor{gray!20}1.00 & \cellcolor{gray!20}1.19 & \cellcolor{gray!20}\textbf{1.40} \\
Qwen3-32B   & 1.00 & 1.06 & 1.26 & 1.00 & 1.15 & 1.34 & 1.00 & 1.10 & 1.27 & 1.00 & 1.08 & 1.27 & 1.00 & 1.05 & 1.26 & \cellcolor{gray!20}1.00 & \cellcolor{gray!20}1.09 & \cellcolor{gray!20}\textbf{1.28} \\
\midrule
\multicolumn{19}{c}{\textbf{InfLLM-v2}} \\
\midrule
GLM-4       & 1.02 & 1.34 & 1.64 & 1.00 & 1.19 & 1.38 & 1.01 & 1.34 & 1.58 & 1.02 & 1.34 & 1.62 & 1.01 & 1.31 & 1.59 & \cellcolor{gray!20}1.01 & \cellcolor{gray!20}1.30 & \cellcolor{gray!20}\textbf{1.56} \\
GLM-Z1      & 1.03 & 1.35 & 1.61 & 1.00 & 1.17 & 1.35 & 1.02 & 1.37 & 1.61 & 1.04 & 1.39 & 1.64 & 1.00 & 1.28 & 1.55 & \cellcolor{gray!20}1.02 & \cellcolor{gray!20}1.31 & \cellcolor{gray!20}\textbf{1.55} \\
DeepSeek-R1 & 1.00 & 1.08 & 1.30 & 1.04 & 1.28 & 1.48 & 1.03 & 1.37 & 1.61 & 1.00 & 1.29 & 1.52 & 1.00 & 1.06 & 1.28 & \cellcolor{gray!20}1.01 & \cellcolor{gray!20}1.22 & \cellcolor{gray!20}\textbf{1.44} \\
Llama3      & 1.00 & 1.12 & 1.32 & 1.04 & 1.26 & 1.45 & 1.00 & 1.31 & 1.55 & 1.00 & 1.30 & 1.55 & 1.00 & 1.07 & 1.30 & \cellcolor{gray!20}1.01 & \cellcolor{gray!20}1.21 & \cellcolor{gray!20}\textbf{1.43} \\
Qwen3-14B   & 1.00 & 1.31 & 1.56 & 1.03 & 1.23 & 1.43 & 1.00 & 1.28 & 1.51 & 1.00 & 1.29 & 1.53 & 1.00 & 1.27 & 1.53 & \cellcolor{gray!20}1.01 & \cellcolor{gray!20}1.28 & \cellcolor{gray!20}\textbf{1.51} \\
Qwen3-32B   & 1.00 & 1.11 & 1.35 & 1.00 & 1.11 & 1.27 & 1.00 & 1.09 & 1.29 & 1.00 & 1.12 & 1.32 & 1.00 & 1.13 & 1.35 & \cellcolor{gray!20}1.00 & \cellcolor{gray!20}1.11 & \cellcolor{gray!20}\textbf{1.32} \\
\bottomrule
\end{tabular}
\end{table*}

\end{document}

%% file: tex/introduction.tex
\section{Introduction}
\label{s:introduction}

Large language models (LLMs) are increasingly deployed for long-context workloads such as deep research~\cite{deepresearcher-1,deepresearcher-2}, multi-step reasoning~\cite{reasoning-1,reasoning-2}, and tool-using agents~\cite{agent-1,agent-2}. 
However, decoding over long contexts remains expensive because the runtime has to access and compute over a growing KV cache for every generated token.

Dynamic sparse attention (DSA) offers a promising remedy by selecting, at each decoding step, only the top-$K$ Key-Value (KV) blocks most relevant to the current query~\cite{quest,infllm,infllm-v2,nsa}. 
By adapting the sparsity pattern to each query, DSA preserves the most relevant context while substantially reducing attention computation, making it an attractive approach for efficient LLM decoding.

\begin{figure}
    \centering
    \includegraphics[width=1\linewidth]{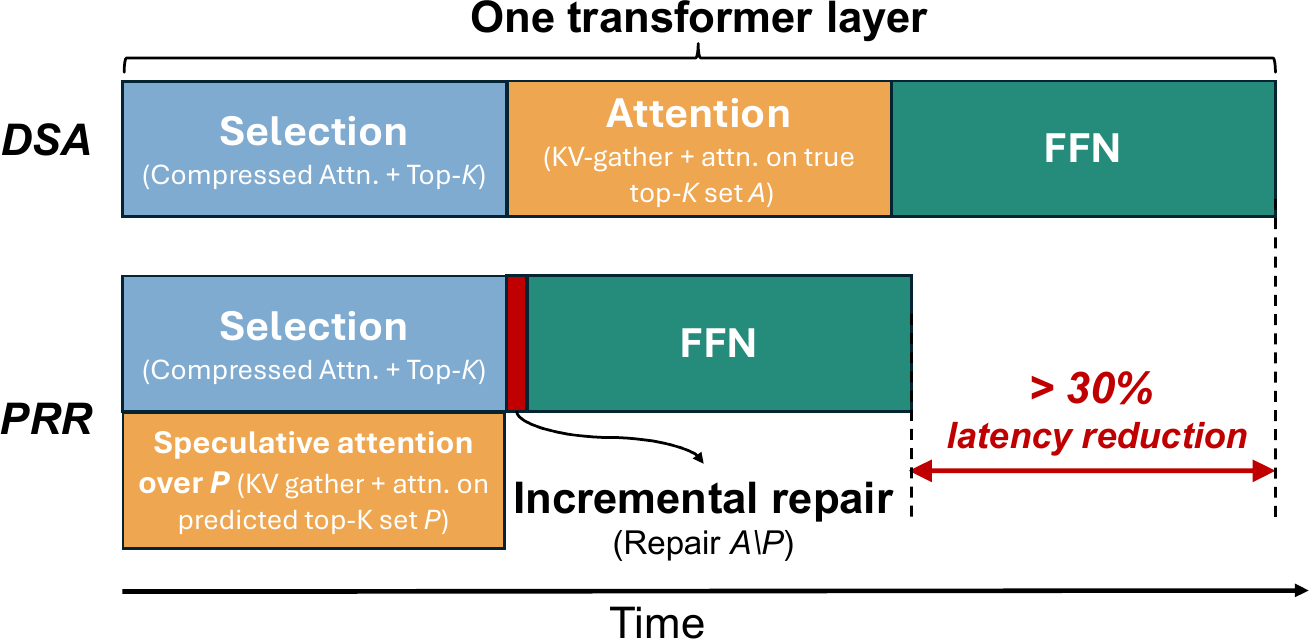}
    \vspace{-7mm}
    \caption{
Standard DSA serializes selection, attention over the true top-$K$ block set $A$, and FFN execution.
\systemname predicts a block set $P$ and executes speculative attention over $P$ in parallel with selection.
After the true set $A$ is known, \systemname incrementally repairs only the missed blocks $A \setminus P$ before FFN execution, saving over 30\% latency for each transformer layer.}
    \label{fig:overview}
    \vspace{-5mm}
\end{figure}

While DSA reduces dense attention arithmetic, it introduces a new bottleneck on the decoding critical path. 
As shown in Figure~\ref{fig:overview}, each decoding step (i.e., a transformer layer) consists of {\it selection}, {\it attention}, and {\it FFN stages}. 
Before attention can run, DSA must first run compressed attention and identify the top-$K$ KV blocks relevant to the current query (within the Selection phase). 
Because these block identities are unknown until selection completes, attention is strictly serialized after selection. 
As context length grows, this \emph{selection-to-attention} dependency becomes increasingly costly, with selection accounting for up to 41\% of per-token generation latency (§\ref{s:motivation}).

\noindent \textbf{Dynamic, Yet Predictable}. 
Although top-$K$ block selections are computed online in dynamic sparse attention, prior work has observed that such selections often exhibit temporal stability across decoding steps~\cite{infinigen,levy2026dynamic_pattern}. 
We find that this property also holds broadly for the DSA methods and various workloads. 
In our measurements, about 68\% of selected blocks are reused across consecutive decoding steps (\S\ref{ss:obs_speculation_opportunity}).


This temporal locality opens an opportunity to \emph{overlap block selection with attention execution}, as illustrated in Figure~\ref{fig:overview}. 
Rather than waiting for the current top-$K$ set to be finalized, the runtime can anticipate likely selected blocks, migrate their KV-cache entries, and compute attention speculatively while block selection is still in flight. 
When the prediction overlaps with the true selected set, the corresponding attention work is removed from the post-selection critical path, thereby reducing token generation latency. 
Realizing this opportunity, however, requires addressing two challenges.

\noindent $\bullet$ {\it First, temporal locality alone is not sufficient}. Blindly reusing the previous step's top-$K$ block indices for the current step yields only a modest hit rate. 
Since each missed block must still be incorporated into the final attention output, a low hit rate leaves much of the sparse attention work on the critical path and limits the benefit of speculation.

\noindent $\bullet$ {\it Second, speculation must preserve the DSA result despite inevitable mispredictions}. 
The final attention output should be computed over the true DSA-selected block set.
However, existing inference engines such as vLLM~\cite{vllm} and SGLang~\cite{sglang}, as well as attention kernels such as FlashAttention~\cite{dao2023flashattention2}, do not expose an interface for incrementally incorporating missed blocks into a partially computed attention result. 
As a result, correcting a miss typically requires recomputing attention from scratch over the true selected set, forfeiting much of the latency benefit of speculative attention computation.


In this paper, we present \systemname (predict, reuse, and repair), a correctness-preserving speculative attention runtime for DSA that addresses both challenges.
To raise prediction accuracy beyond what prior-step reuse offers, we introduce a lightweight, exponential moving average (EMA)-based predictor that tracks the temporal trajectory of per-block importance scores and \emph{anticipates}, rather than lags, the upcoming top-$K$ set. Its smoothing hyperparameters are calibrated per-prompt using importance scores already produced during prefill, adding negligible model execution cost.

To overlap attention computation with block selection, \systemname executes speculative attention over a predicted block set $P$ in parallel with the current top-$K$ selection.
The size of $P$ is controlled by a profiled dynamic budget that keeps speculative execution off the critical path, i.e., it won't delay FFN.
Once the true top-$K$ block set $A$ becomes available, blocks in $P\cap A$ have already been processed, while missed blocks in $A\setminus P$ are incorporated through incremental attention repair.
Blocks in $P\setminus A$ are retained as additional low-priority context.


To make repair efficient, \systemname implements a customized CUDA kernel based on FlashAttention's online-softmax recurrence that performs \emph{incremental} attention repair, as depicted in Figure~\ref{fig:overview}. 
Given the speculative output and its running log-sum-exp statistics, the kernel attends only to the missed blocks and merges their contribution into the existing accumulator.
This incurs no accuracy loss since all blocks in $A$ are included, while bounding repair work by $|A \setminus P|$ rather than $|A|$.

With the above design, \systemname preserves the semantics of standard DSA: although it speculates on likely top-$K$ blocks, the final attention output always covers the true DSA-selected blocks after repairing. 
As a result, \systemname maintains the same accuracy as the original DSA implementation.
Across LongBench~\cite{longbench}, InfiniteBench~\cite{infinitebench}, RULER~\cite{ruler}, AIME~\cite{hf_aime_2024}, and MATH500~\cite{hf_math500}, \systemname achieves per-token decoding speedup by $1.42\times$ over Quest and $1.56\times$ over InfLLM-v2 across multiple LLMs, while maintaining the same downstream task accuracy.
Our contributions are listed below:
\squishlist{}
    \item We identify the {\it selection-to-attention} dependency as a new critical-path bottleneck in dynamic sparse attention mechanisms.
    \item We design \systemname, a runtime that overlaps speculative attention over predicted blocks with selection, reuses correctly predicted block states, and repairs only missed blocks.
    \item We implement an online-softmax-based incremental repair kernel that ensures full top-$K$ coverage as standard DSA and bounds post-selection work by the number of missed blocks.
\squishend{}

%% file: tex/motivation.tex
\section{Motivation}
\label{s:motivation}

In this section, we first show that DSA shifts decoding cost to selection-attention dependency (\S\ref{ss:obs_selection_to_gather}). 
We then show that sparse indices exhibit strong predictability, and that DSA decoding leaves enough idle GPU resources to turn this predictability into an opportunity for speculative attention (\S\ref{ss:obs_speculation_opportunity}).

\noindent \textbf{Setup}. 
We use GLM-4-9B~\cite{glm-4} to evaluate two representative DSA methods, Quest~\cite{quest} and InfLLM-V2~\cite{infllm-v2}, across long-sequence benchmarks and a range of context lengths. 
We run these studies on NVIDIA H100 GPUs, with FlashInfer's \texttt{BlockSparseAttention}~\cite{ye2025flashinfer} serving as the optimized sparse-attention backend. 

\begin{figure}[t]
    \centering
    \includegraphics[width=1\linewidth]{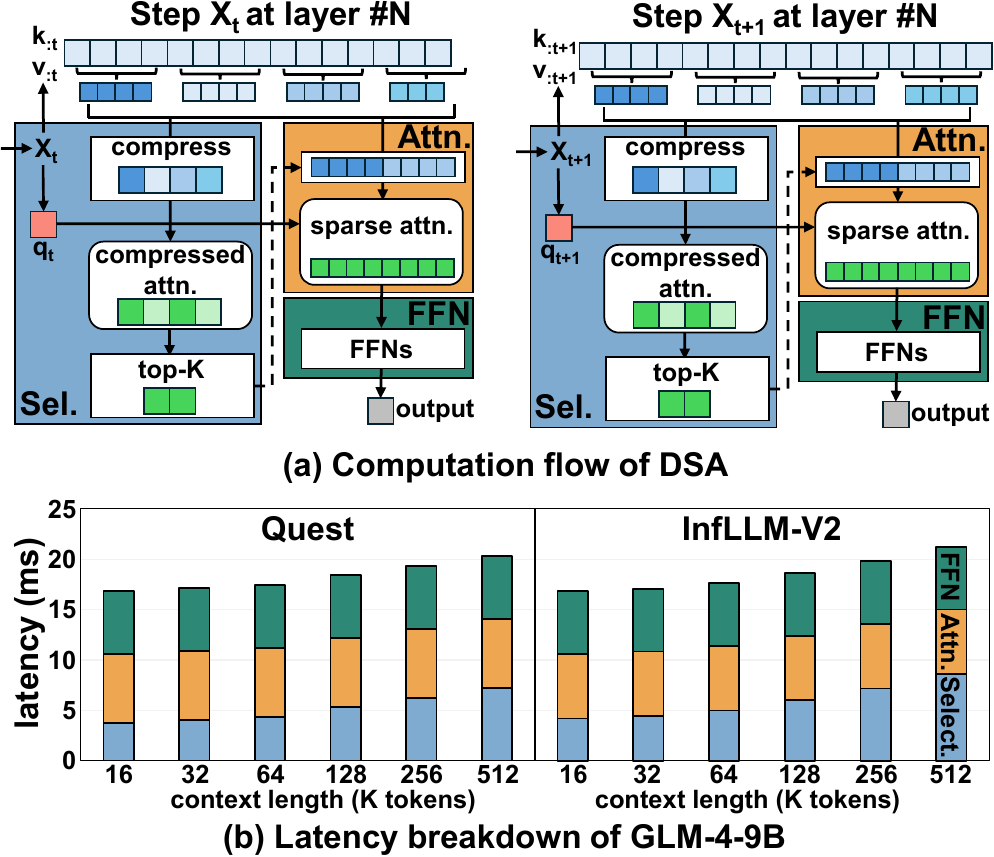}
    \vspace{-5mm}
    \caption{(a) Three stages in DSA and temporal similarity in two consecutive top-$K$ selections, and (b) latency breakdown of GLM-4-9B with Quest and InfLLM-V2 under various context lengths.}
    \vspace{-5mm}
    \label{fig:latency_breakdown}
\end{figure}

\subsection{The Selection-to-Attention Dependency}
\label{ss:obs_selection_to_gather}

\begin{observationbox}{Observation One:}
DSA significantly reduces attention computation, but exposes a serialized selection-to-attention dependency on the decoding critical path.
\end{observationbox}
Figure~\ref{fig:latency_breakdown}(a) shows the computation flow of DSAs. At each layer, the runtime partitions KV tokens into blocks, compresses each block (e.g., mean), and computes compressed attention to pick the top-$K$ blocks for the current step. Only then can the runtime fetch the selected KV blocks and execute (sparse) attention, followed by the FFN. This creates a strict selection-to-attention dependency.


Figure~\ref{fig:latency_breakdown}(b) shows profiled latency of three stages: selection, attention, and FFN. The serialized selection-to-attention path consumes about 60\% of decoding latency at 16K tokens and rises to ~71\% at 512K.
Selection alone grows from ~4\,ms to ~8\,ms over the same range. Since attention is blocked until top-$K$  selection finishes, this growing selection cost directly delays the attention stage and exposes a substantial critical-path bottleneck.

\subsection{The Opportunity for Speculation}
\label{ss:obs_speculation_opportunity}


Prior works~\cite{infinigen,takbir2026flexicache} have shown that the same tokens tend to be selected or to receive high attention scores across consecutive decoding steps.
Motivated by this, we measure how many blocks are repeatedly selected in consecutive decoding steps on five representative long-context benchmarks, including LongBench~\cite{longbench}, InfiniteBench~\cite{infinitebench}, AIME~\cite{hf_aime_2024}, MATH500~\cite{hf_math500}, and RULER~\cite{ruler}.

\begin{observationbox}{Observation Two:} The top-$K$ blocks in DSA show strong locality across adjacent decoding steps, which leaves an opportunity to speculative attention.
\end{observationbox}


\begin{figure}[t]     
  \centering
  \includegraphics[width=\linewidth]{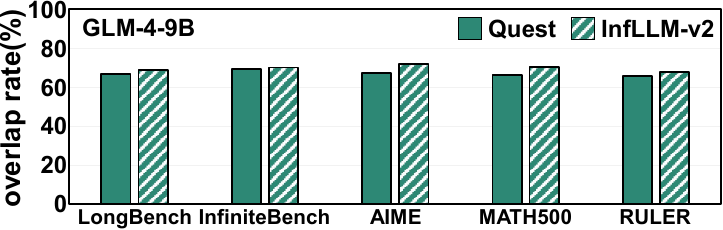}
  \caption{Average overlap rate of blocks in consecutive selections of GLM-4-9B under various benchmarks.}
  \label{fig:temporal_locality}
  \vspace{-5mm}             
\end{figure}

Figure~\ref{fig:temporal_locality} shows that temporal block selection locality is not confined to a particular benchmark or DSA method. 
Across LongBench, InfiniteBench, AIME, MATH500, and RULER, both Quest and InfLLM-V2 consistently reselect more than 65\% of blocks across consecutive decoding steps, with an average overlap of about 68\%\footnote{We have similar observations on other LLMs in~\ref{ss:appdix_temporal_locality}.}.

This stability across diverse long-context and reasoning workloads suggests that DSA selections are dynamic but sufficiently predictable for speculation.
Instead of waiting for the current top-$K$ selection to complete, the runtime can predict likely selected blocks and begin executing attention over them while selection is still in progress, as shown in Figure~\ref{fig:overview}.
Correctly predicted blocks remove their KV gathering and attention computation from the critical path, while partial predictions leave only the missed blocks to be processed after selection completes. 
This motivates \systemname's speculative attention-and-repair execution model.




While temporal locality makes speculation possible, it also raises a new question: {\it does the GPU have enough spare memory-bandwidth and compute resources to execute speculative KV gathering and attention computation without slowing down the normal DSA pipeline?}
To answer this question, we profile GPU resource utilization for DSAs.

\begin{observationbox}{Observation Three:}
Long-context DSA decoding leaves sufficient idle GPU resources to support speculative KV gathering and attention computation.
\end{observationbox}

\begin{figure}[]     
  \centering
  \includegraphics[width=\linewidth]{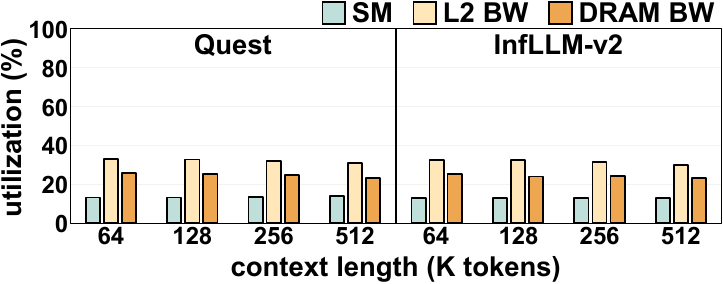}
  \caption{SM utilization, L2 bandwidth, and DRAM bandwidth profiled for GLM-4-9B paired with Quest and InfLLM-v2 across varying context lengths.}
  \label{fig:utilization}
  \vspace{-5mm}             
\end{figure}

We measure the utilization of streaming multiprocessors (SMs), L2 bandwidth, and DRAM bandwidth during decoding across a range of context lengths, up to 512K tokens.
As shown in Figure~\ref{fig:utilization}, even at a 512K context length, SM, L2-bandwidth, and DRAM-bandwidth utilization all remain below 40\%.
This substantial headroom indicates that ample idle resources are available to precompute the sparse attention speculatively, based on estimated block selections, without contending with the compressed attention on the critical path.
The same trend holds at larger batch sizes, as shown in~\ref{ss:appdix_low_util}.

\section{The Design Space}
\label{ss:design_space}






These observations suggest a design that, given idle bandwidth and compute, predicts likely blocks, prefetches their KV, and overlaps their attention computation with top-$K$ block selection. This opens two interleaved design spaces to explore:

\noindent\textbf{Design Space One: Maximize speculative attention accuracy.}
Let $A$ be the set of blocks chosen by the ground-truth top-$K$ selection in the current round, and $P$ denote the predicted blocks. 
We categorize the prediction errors into two distinct types:

\squishlist{}
    \item {\it Missed Blocks ($M = A \setminus P$)}: these blocks lie on the critical path: every block in $M$ must be migrated and attended to after the true selection completes, and the partial attention output must be corrected accordingly.
    \item {\it Wasted Blocks ($N = P \setminus A$)}: these blocks consume extra bandwidth and computing resources, that waste the precious resources. But they don't extend the critical path.
\squishend{}
\begin{figure}[t]
    \centering
    \includegraphics[width=1\linewidth]{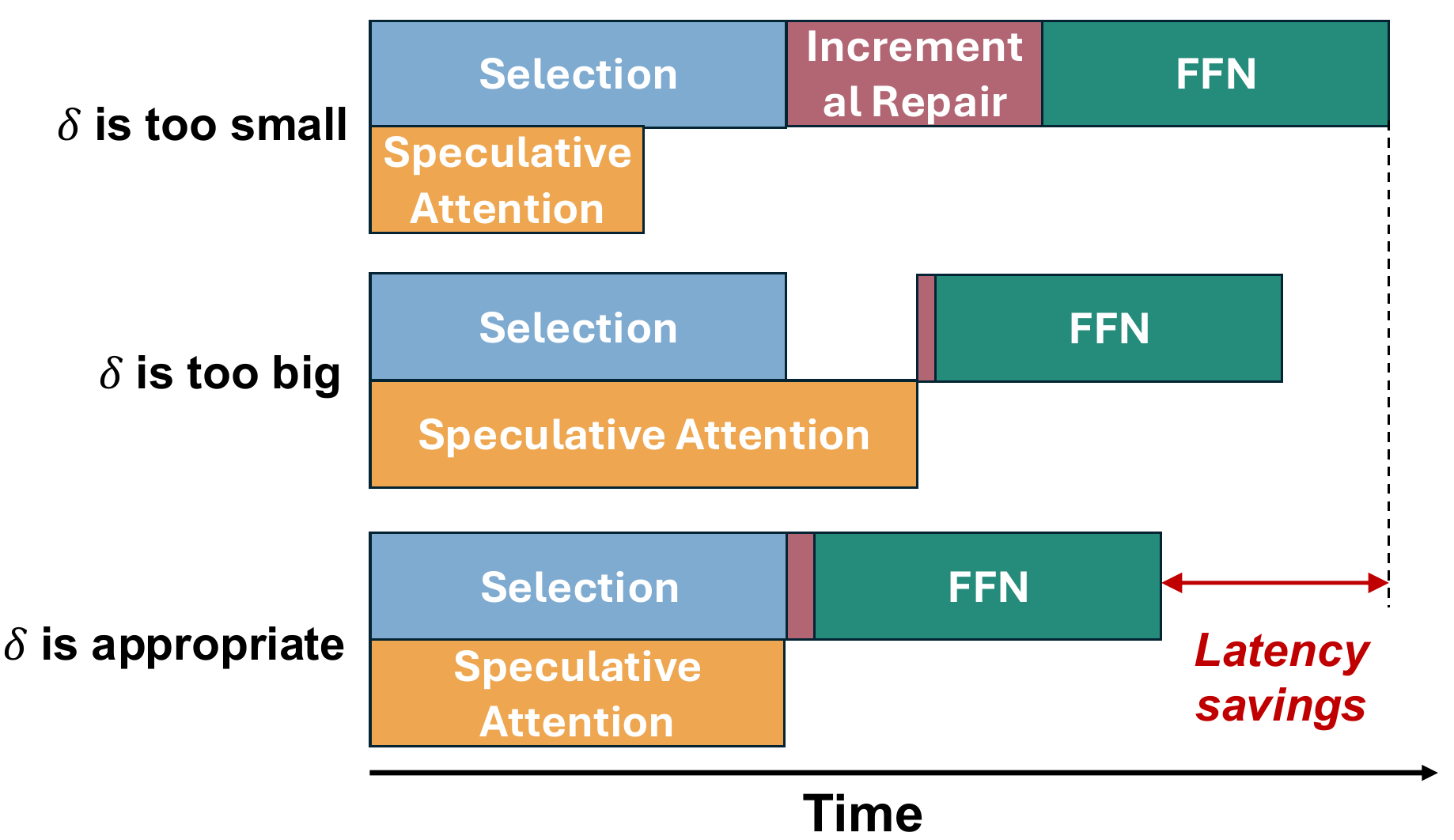}
    \caption{Impact of the speculation budget ratio $\delta$ on PRR's execution timeline.
(top): When $\delta$ is too small, the predicted set covers too few true top-$K$ blocks, leaving a large incremental repair stage on the critical path.
(middle): When $\delta$ is too large, speculative attention itself exceeds the top-$K$ selection window and delays the FFN.
(bottom): An appropriate $\delta$ balances coverage and speculative cost, allowing speculative attention to overlap with top-$K$ selection without delaying each other while leaving only a small repair stage before FFN.
}
\vspace{-6mm}
    \label{fig:alpha}
\end{figure}
The two costs are therefore asymmetric, and the dominant objective is to minimize $|M|$.
We cast this as a single-objective problem with a budget constraint:
\begin{equation}
\min \; |M| = |A \setminus P| \quad \text{s.t.} \quad |P| \leq \delta |A|
\label{eq:goal}
\end{equation}
where $\delta$ is the budget ratio that bounds how many blocks the speculator may fetch. 

A larger $\delta$ admits more candidates and likely reduces $|M|$, but increases bandwidth pressure and speculative attention computation cost.
However, if too many blocks are included, speculative attention may take longer than block selection, extending the critical path and negating the latency gains from speculation, as shown in Figure~\ref{fig:alpha}.


\noindent $\Rightarrow$ To shrink $|M|$, we replace the naive ``reuse previous top-$K$'' heuristic with an EMA-based predictor that tracks block-importance scores across past decoding steps and predicts a top-$K$ set $P$ likely to be selected in the current round (\S\ref{ss:design_ema}).
Furthermore, rather than fixing $\delta$ statically, we dynamically choose the speculation budget using offline profiles of block-selection latency and speculative-attention latency across context lengths, ensuring that speculative attention remains aligned with the selection stage and does not extend the critical path.


\noindent\textbf{Design Space Two: Enable incremental attention repair.}
Even with a high-quality predictor, mispredictions are unavoidable, i.e., $|M| > 0$.
Once the true top-$K$ set $A$ is known, the runtime must reconcile the speculative attention result with the true selected blocks.
A simple fallback is to discard the speculative result and recompute attention over $A$ from scratch.
This preserves correctness, but it puts the attention computation back on the critical path, effectively reverting to standard DSA and leaving little to no latency benefit.

\noindent $\Rightarrow$ To avoid this fallback, we propose a customized attention kernel, built on top of FlashAttention, that supports \emph{incremental repair}. 
Given the attention state over the predicted set $P$, including the output and running softmax statistics, the kernel attends only to the missed blocks $M = A \setminus P$ and merges them into the existing accumulator via log-sum-exp rescaling. 
This guarantees the full coverage over the true selected set $A$ as standard DSA, while reducing post-selection work from $|A|$ blocks to $|M|$ blocks. 
Thus, the latency benefit scales with prediction coverage instead of being lost whenever the prediction is imperfect.

%% file: tex/design.tex
\section{Design}
\label{s:design}

To explore these design spaces, we propose \systemname, a {\it speculation-and-repair} runtime that predicts likely blocks (\S\ref{ss:design_ema}) and overlaps sparse attention with block selection. The speculator is calibrated online with negligible delay (\S\ref{ss:design_hyperparams_determine}).
\systemname then incrementally incorporates missed blocks once the true top-$K$ set $A$ is available (\S\ref{ss:attention_correctness}), and dynamically sizes the speculative set $P$ to maximize coverage of $A$ without extending the critical path (\S\ref{ss:budget}).

\subsection{The Lightweight, EMA-based Predictor}
\label{ss:design_ema}

\systemname requires a predictor that is accurate enough to cover the true top-$K$ blocks as much as possible, yet lightweight enough to run off the critical path. 
A neural network-based predictor is promising for higher accuracy, but it would introduce additional model execution and memory traffic, which can offset the latency saved by speculation. 
We therefore choose a training-free, ultra-lightweight exponential moving average (EMA)-based predictor that operates directly on the block-importance scores already produced by DSA.

We denote by $IS_i^t$ the true importance score of block $i$ at decoding step $t$, produced by the DSA's top-$K$ block selection stage. 
Before this stage runs at step $t$, \systemname predicts the score of each block as $\widehat{IS}_i^t$ using only historical scores observed up to step $t{-}1$. 
The predicted scores are used to select a speculative block set $P$, with the goal of maximizing coverage of the true top-$K$ $A$ set under a bounded speculation budget.



\noindent\textbf{State and initialization.}
For each block $i$, the predictor maintains two state variables after each step $t$: a smoothed level $\ell_i^t$ and a trend estimate $v_i^t$. 
When block $i$ first enters the KV cache at step $\tau$, we initialize
$\ell_i^\tau = IS_i^\tau$ and $v_i^\tau = 0$ because no prior trend is available.

\noindent\textbf{Prediction.}
At the beginning of step $t$, before the current DSA selection stage produces $IS_i^t$, the predictor extrapolates from the previous state:
\begin{equation}
\widehat{IS}_i^t = \ell_i^{t-1} + \gamma v_i^{t-1},
\label{eq:ema_predict}
\end{equation}
where $\gamma \in [0,1]$ controls how aggressively the predictor extrapolates the recent trend.

\noindent\textbf{Update.}
After the DSA selection stage at step $t$ produces the true score $IS_i^t$, the predictor updates its state using the following equations:
\begin{equation}
\begin{split}
v_i^t &= \beta \bigl(IS_i^t - \ell_i^{t-1}\bigr) + (1-\beta)v_i^{t-1}, \\
\ell_i^t &= \alpha IS_i^t + (1-\alpha)\ell_i^{t-1}.
\end{split}
\label{eq:ema_update}
\end{equation}
Here, $\alpha \in (0,1]$ controls how quickly the level follows newly observed scores, while $\beta \in (0,1]$ controls how quickly the trend estimate adapts to score changes.

\subsection{Online Predictor Calibration}
\label{ss:design_hyperparams_determine}

The EMA predictor is lightweight, but its prediction accuracy depends on the smoothing hyperparameters $(\alpha,\beta,\gamma)$.
These parameters determine whether the predictor tracks block-importance dynamics smoothly or reacts quickly to recent changes. 
A single fixed $(\alpha,\beta,\gamma)$ setting would not be effective because prompt-specific factors such as task type and context length can substantially change the stability of the top-$K$ trajectory. 

\begin{figure}[t]
    \centering
    \includegraphics[width=1\linewidth]{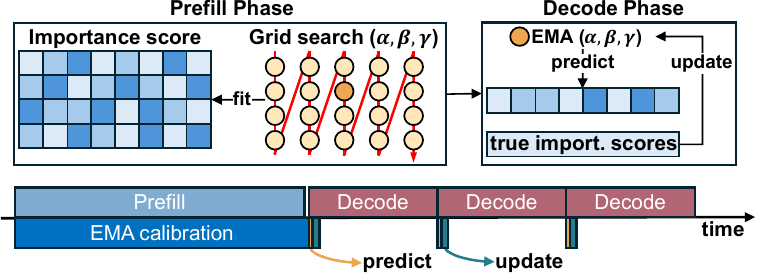}
    \caption{During prefill phase, \systemname calibrates EMA over importance scores with grid search. During decode phase, EMA predicts block importance scores and updates from true scores produced by block selection.}
    \vspace{-5mm}
    \label{fig:ema_calibration}
\end{figure}

\systemname instead uses the prefill phase to calibrate these parameters per prompt, as depicted in Figure~\ref{fig:ema_calibration}. Because DSA already computes compressed-attention importance scores for each prefill token, prefill provides a ready-made trajectory of block scores at no additional model-execution cost. 
\systemname searches over a small $(\alpha,\beta,\gamma)$ candidate grid on this trajectory and selects the setting that best predicts prefill top-$K$ selections, yielding a prompt-adaptive predictor before decoding starts.
We elaborate on the design below.


\noindent\textbf{Prefill trajectory.} Let $T_p$ denote the number of prefill tokens.
The prefill phase produces the matrix $IS \in \mathbb{R}^{T_p \times N_p}$, where row $\phi$ is the compressed-attention score vector at prefill token $\phi$ over the $N_p$ dynamic blocks formed during prefill.
This matrix is fully available before the first decoding step.


\noindent\textbf{Calibration objective.} Given a candidate hyperparameter setting $\theta = (\alpha, \beta, \gamma)$, we simulate the predictor over $IS$ from prefill, producing a predicted selection set $P_\phi(\theta)$ for each prefill token $\phi$, and measure the resulting score-weighted hit rate:
\begin{equation}
  H\bigl(\theta \,;\, IS\bigr) \;=\;
  \frac{1}{T_p}\sum_{\phi}
    \frac{\sum_{b \in P_\phi(\theta) \cap A_\phi} IS_{\phi,b}}{\sum_{b \in A_\phi} IS_{\phi,b}}
  \label{eq:cal-obj}
\end{equation}
where $A_{\phi}$ is the ground-truth top-$K$ block selection set at prefill token $\phi$ and $IS_{\phi,b}$ is the importance score of block $b$ at token $\phi$.
The per-prompt hyperparameter choice is formulated as:
\begin{equation}
    \theta^{\star} = \arg\max_\theta H\bigl(\theta; IS\bigr)
\end{equation}
\noindent\textbf{Search procedure.} We perform grid search rather than gradient-based solutions for two reasons.
First, gradient-based hyperparameter optimization of a set-valued objective (top-$K$ membership) requires a differentiable surrogate for argsort, which introduces its own approximation error and implementation dependency.
Second, the search space is small, so a direct search is both simpler and sufficient.

We set the search grid in our system as follows: $\alpha \in [0.2, 0.8]$ with step $0.2$, $\beta \in [0.1, 0.5]$ with step $0.1$, and $\gamma \in [0, 0.75]$ with step $0.25$, yielding $|\Theta| = 80$ candidates per prompt. 
The search overlaps with the prefill phase and adds minimal extra latency (0.06\,ms) to the critical path (\ref{ss:design_search_prediction_cost}).

\subsection{Incremental Repair via Online Softmax}
\label{ss:attention_correctness}



Since mispredictions are unavoidable (i.e., $M = A \setminus P \neq \emptyset$) and softmax is sensitive to the maximum logit, ignoring $M$ produces an inexact output, and the error compounds across decoding steps. 
We propose a repair phase whose cost scales only with $|M|$, built on FlashAttention's online-softmax recurrence~\cite{dao2023flashattention2}.
Our customized CUDA kernel enables incremental attention repair: a capability missing from existing inference engines~\cite{vllm,sglang} and backend kernels~\cite{dao2023flashattention2, ye2025flashinfer}.

\noindent\textbf{Instrumented speculative kernel.} We modify the FlashAttention forward to write not only the attention output $O$ but also the denominator $\ell$ and the running maximum $m$ to HBM.
The change is surgical: the inner-loop arithmetic is unchanged, and the additional store amounts to two scalars per head, so the kernel retains the memory-access pattern and occupancy of the original forward.
The extra bandwidth is negligible relative to KV traffic that already dominates the pass.

\noindent\textbf{Repair kernel.} Given $(O^{\text{spec}}, \ell^{\text{spec}}, m^{\text{spec}})$ from the speculative sparse attention and the KV entries of $M$, a second kernel applies the recurrence in Equations~\eqref{eq:online-max}--\eqref{eq:online-out} once per missed block:
\begin{align}
   m^{(t+1)}   &= \max(m^{(t)}, \tilde{m}) \label{eq:online-max}\\
  \ell^{(t+1)} &= e^{m^{(t)} - m^{(t+1)}}\,\ell^{(t)} + e^{\tilde{m} - m^{(t+1)}}\,\tilde{\ell} \label{eq:online-sum}\\
  O^{(t+1)}    &= \frac{e^{m^{(t)} - m^{(t+1)}}\,\ell^{(t)}}{\ell^{(t+1)}}\, O^{(t)} \notag\\
               &+ \frac{e^{\tilde{m} - m^{(t+1)}}} {\ell^{(t+1)}} \sum_{j} e^{S_{t+1,j} - \tilde{m}}\, V_{t+1,j}  \label{eq:online-out}
\end{align}

Each query is handled independently. The missed blocks are streamed through on-chip memory in the same tiled fashion as the FlashAttention forward, so no intermediate score matrix is materialized.
The kernel reads $O^{\text{spec}}$, $\ell^{\text{spec}}$, $m^{\text{spec}}$, and the missed $(K, V)$ tiles once, and writes the corrected $(O, \ell, m)$ once.
Both FLOPs and memory traffic scale solely with $|M|$.

\subsection{Critical-Path-Aware Speculation Budget} 
\label{ss:budget}

Recall from Figure~\ref{fig:alpha} that a larger $\delta$ allows \systemname to include more predicted blocks in speculative attention, which increases the likelihood that $P$ covers the true selected set $A$ and reduces the missed blocks $A \setminus P$.
However, it also increases speculative KV movement and speculative attention computation, which may extend the critical path if speculative attention takes longer than selection.

\systemname therefore chooses $\delta$ dynamically rather than fixing it globally.
For each model-hardware-DSA configuration, we perform a one-time offline profiling sweep that measures selection latency and speculative-attention latency under different context lengths and budget ratios.
Because the model, hardware backend, DSA method, block size, and DSA top-$K$ are fixed, these latencies are primarily determined by context length and $\delta$.
\systemname stores the profiling results in a lightweight lookup table indexed by context length, and at runtime selects the largest safe $\delta$ whose speculative attention latency fits within the current selection window.
This ensures that speculative execution remains aligned with block selection and does not extend the critical path.
In our implementation, we profile context lengths of 4K, 8K, 16K, 32K, and 64K, with details and profiling overhead reported in \ref{ss:appdix_profiling_overheads}.
Note that here we use five context lengths as an example, one may profile in a much finer-grained manner to achieve more latency reduction.

%% file: tex/evaluation.tex
\section{Evaluation}
\label{s:eval}

We describe the system evaluation in this section.

\subsection{Experiment Setups}
\label{ss:experiment_setup}
\noindent\textbf{Evaluation Dataset}. We evaluate \systemname on five benchmarks covering long-context understanding (e.g., summarization) and complex reasoning (e,g., math). Long-context tasks are assessed using LongBench~\cite{longbench}, InfiniteBench~\cite{infinitebench}, and RULER~\cite{ruler}, while reasoning capability is measured on AIME~\cite{hf_aime_2024} and MATH500~\cite{hf_math500}.

\noindent\textbf{Models. }We conduct experiments on six LLMs: GLM-4-9B-1M~\cite{glm-4}, GLM-Z1-9B~\cite{glm-z1}, DeepSeek-R1-8B~\cite{deepseek-r1}, Llama3-8B-1M~\cite{llama3-8b-1m}, Qwen3-14B~\cite{qwen3}, and Qwen3-32B~\cite{qwen3}.

\noindent\textbf{DSAs. }We consider two plug-and-play DSA mechanisms: Quest~\cite{quest} and InfLLM-v2~\cite{infllm-v2}. NSA~\cite{nsa} is excluded as it requires training from scratch for its feature extraction and gating components.

\noindent\textbf{Testbed. } Experiments are run on H100 GPUs with CUDA 12.8, with a tensor parallelism degree of 2.

\subsection{Accelerate Evaluation}

\begin{table}[]
\centering
\footnotesize
\setlength{\tabcolsep}{0.6pt} 
\caption{Decoding speedup of \systemname relative to the serial DSA execution ($1.00\times$). Higher is better.}
\label{tab:main_result}
\begin{tabular}{lcccccc}
\toprule
\textbf{Model} & \textbf{\shortstack{Long\\Bench}} & \textbf{\shortstack{Infinite\\Bench}} & \textbf{RULER} & \textbf{AIME} & \textbf{\shortstack{MATH\\500}} & \cellcolor{gray!20}\textbf{Avg.} \\
\midrule
\multicolumn{7}{c}{\textbf{Quest}} \\
\midrule
GLM-4       & 1.50$\times$ & 1.40$\times$ & 1.45$\times$ & 1.30$\times$ & 1.42$\times$ & \cellcolor{gray!20}\textbf{1.42$\times$} \\ \hline
GLM-Z1      & 1.48$\times$ & 1.39$\times$ & 1.43$\times$ & 1.33$\times$ & 1.45$\times$ & \cellcolor{gray!20}\textbf{1.42$\times$} \\ \hline
DeepSeek-R1 & 1.39$\times$ & 1.42$\times$ & 1.47$\times$ & 1.35$\times$ & 1.39$\times$ & \cellcolor{gray!20}\textbf{1.41$\times$} \\ \hline
Llama3      & 1.36$\times$ & 1.40$\times$ & 1.41$\times$ & 1.39$\times$ & 1.34$\times$ & \cellcolor{gray!20}\textbf{1.38$\times$} \\ \hline
Qwen3-14B   & 1.44$\times$ & 1.27$\times$ & 1.41$\times$ & 1.44$\times$ & 1.45$\times$ & \cellcolor{gray!20}\textbf{1.40$\times$} \\ \hline
Qwen3-32B   & 1.26$\times$ & 1.34$\times$ & 1.27$\times$ & 1.27$\times$ & 1.26$\times$ & \cellcolor{gray!20}\textbf{1.28$\times$} \\ 
\hline
\midrule
\multicolumn{7}{c}{\textbf{InfLLM-v2}} \\
\midrule
GLM-4       & 1.64$\times$ & 1.38$\times$ & 1.58$\times$ & 1.62$\times$ & 1.59$\times$ & \cellcolor{gray!20}\textbf{1.56$\times$} \\ \hline
GLM-Z1      & 1.61$\times$ & 1.35$\times$ & 1.61$\times$ & 1.64$\times$ & 1.55$\times$ & \cellcolor{gray!20}\textbf{1.55$\times$} \\ \hline
DeepSeek-R1 & 1.30$\times$ & 1.48$\times$ & 1.61$\times$ & 1.52$\times$ & 1.28$\times$ & \cellcolor{gray!20}\textbf{1.44$\times$} \\ \hline
Llama3      & 1.32$\times$ & 1.45$\times$ & 1.55$\times$ & 1.55$\times$ & 1.30$\times$ & \cellcolor{gray!20}\textbf{1.43$\times$} \\ \hline
Qwen3-14B   & 1.56$\times$ & 1.43$\times$ & 1.51$\times$ & 1.53$\times$ & 1.53$\times$ & \cellcolor{gray!20}\textbf{1.51$\times$} \\ \hline
Qwen3-32B   & 1.35$\times$ & 1.27$\times$ & 1.29$\times$ & 1.32$\times$ & 1.35$\times$ & \cellcolor{gray!20}\textbf{1.31$\times$} \\ 
\hline
\end{tabular}
\end{table}

We measure end-to-end decoding latency across all models and benchmarks, and report the speedup of \systemname over the standard DSA execution in Table~\ref{tab:main_result}. 
As shown, \systemname delivers consistent speedups, achieving an average of $1.35\times$ speedup under Quest and $1.47\times$ speedup under InfLLM-v2.
These gains come from three complementary mechanisms. 
First, the EMA predictor achieves an average top-$K$ overlap rate of roughly 91\%, enabling most of the attention work done during speculation.
Second, our customized sparse attention CUDA kernel substantially outperforms FlashInfer's \texttt{BlockSparseAttention} (\S\ref{ss:kernel_speedup}). 
The headroom it creates lets \systemname include more blocks in speculation, lifting the overlap rate from 91\% to 98\% at no added latency cost (\S\ref{ss:ema_accuracy}). 
Third, when block misses do occur, incremental attention repair patches in the missing blocks rather than recomputing attention from scratch, preserving the latency savings from speculation.

\subsection{Ablation Study}
We quantify the contribution of each design component to the overall latency reduction by progressively enabling three features:
\squishenum{}
\item \textbf{Reuse the top-$K$ blocks in the last step:} Speculatively compute attention by using the block selections from the previous decoding step.
\item \textbf{Incremental attention repair:} Replace FlashInfer's \texttt{BlockSparseAttention} with our customized CUDA kernel, which performs incremental attention repair on missed blocks.
\item \textbf{EMA predictor:} Apply our EMA-based predictor to anticipate the upcoming block selections and speculate over the predicted set.
\squishenumend{}
These features are enabled incrementally from S1 to S3, where S3 represents the standard \systemname configuration.
We use GLM-4-9B paired with InfLLM-v2 as the running example. 
The consistent trend is observed across other LLMs under both Quest and InfLLM-v2 (\ref{ss:appdix_ablation}).

\begin{table}[t]
\centering
\footnotesize
\setlength{\tabcolsep}{10pt}
\caption{Decoding speedup of ablation stages across benchmarks for GLM-4-9B and InfLLM-v2 relative to Serial execution ($1.00\times$). Higher is better.}
\label{tab:ablation}
\begin{tabular}{lccc}
\toprule
\textbf{Benchmark} & \textbf{S1} & \textbf{S2} & \textbf{S3 (\systemname)} \\
\midrule
LongBench     & 1.02$\times$ & 1.34$\times$ & 1.64$\times$ \\ \hline
InfiniteBench & 1.00$\times$ & 1.19$\times$ & 1.38$\times$ \\ \hline
RULER         & 1.01$\times$ & 1.34$\times$ & 1.58$\times$ \\ \hline
AIME          & 1.02$\times$ & 1.34$\times$ & 1.62$\times$ \\ \hline
MATH500       & 1.01$\times$ & 1.31$\times$ & 1.59$\times$ \\ 
\hline
\rowcolor{gray!20}
\textbf{Avg. Speedup} & \textbf{1.01$\times$} & \textbf{1.30$\times$} & \textbf{1.56$\times$} \\
\hline
\end{tabular}%
\end{table}

Results in Table~\ref{tab:ablation} reveal the following. 
Naively enabling speculative attention through S1 yields negligible improvement over standard DSA execution. Although the block selections reused from the previous decoding step overlap with the current true top-$K$ set by roughly 68\% on average, a single missed block triggers a full attention recomputation over the true top-$K$ selections, which offsets the latency savings from speculation.
S2 introduces our customized sparse attention kernel, which allows the run-time to repair the missing $32\%$ blocks, raising the average speedup to $1.30\times$.
S3 (\systemname) attains 98\% overlap rate between the predicted top-$K$ set and the true top-$K$ set, leading to 1.55$\times$ speedup on average.

\subsection{Kernel Speed Comparison}
\label{ss:kernel_speedup}

We compare our customized sparse attention kernel against FlashInfer's \texttt{BlockSparseAttention}~\cite{ye2025flashinfer} on H100 under a GQA configuration with 32 query heads, 8 KV heads, head dimension 128, and a 128K-token context. We sweep the token budget from 1K to 8K and the KV-cache block size from 16 to 256 to reflect realistic sparse attention workloads during decoding.

\begin{table}[t]
\centering
\small
\setlength{\tabcolsep}{2.5pt}
\caption{Speedup of \systemname's sparse kernel over FlashInfer's \texttt{BlockSparseAttention} under various block sizes and token budgets. Higher indicates greater speedup.}
\label{tab:kernel_table}
\begin{tabular}{lcccccc}
\toprule
\textbf{Block Size} & \textbf{1024} & \textbf{2048} & \textbf{4096} & \textbf{6144} & \textbf{8192} & \cellcolor{gray!20}\textbf{Avg.} \\
\midrule
\textbf{16}  & 1.17$\times$ & 1.28$\times$ & 2.00$\times$ & 2.83$\times$ & 3.68$\times$ & \cellcolor{gray!20}\textbf{2.19$\times$} \\ \hline
\textbf{32}  & 1.17$\times$ & 1.38$\times$ & 2.09$\times$ & 2.78$\times$ & 3.69$\times$ & \cellcolor{gray!20}\textbf{2.22$\times$} \\ \hline
\textbf{64}  & 1.17$\times$ & 1.38$\times$ & 2.26$\times$ & 2.91$\times$ & 3.64$\times$ & \cellcolor{gray!20}\textbf{2.27$\times$} \\ \hline
\textbf{128} & 1.17$\times$ & 1.41$\times$ & 2.23$\times$ & 3.00$\times$ & 3.71$\times$ & \cellcolor{gray!20}\textbf{2.31$\times$} \\ \hline
\textbf{256} & 1.03$\times$ & 1.11$\times$ & 2.00$\times$ & 2.75$\times$ & 3.64$\times$ & \cellcolor{gray!20}\textbf{2.11$\times$} \\
\hline
\end{tabular}
\end{table}

As shown in Table~\ref{tab:kernel_table}, our kernel outperforms \texttt{BlockSparseAttention} across every configuration, with an average speedup of $2.22\times$ and a peak of $3.71\times$ at an 8K budget. 
At 1K tokens, the speedup is modest ($1.03{-}1.17\times$) because absolute latencies are small and fixed overheads dominate, whereas at 8K the bandwidth-saturation advantages compound, yielding $3.64{-}3.71\times$.
Across block sizes, performance is stable; block size 128 yields the highest average speedup ($2.31\times$) and 256 the lowest ($2.11\times$), since the largest tiles approach the dense regime where block-table indirection becomes pure overhead for \texttt{BlockSparseAttention}.

Beyond raw speedup, our kernel exposes an interface for incremental attention repair that patches missed blocks into an existing attention output rather than recomputing from scratch, which is the key capability underlying \systemname's end-to-end gains.

\subsection{EMA Hit Rate}
\label{ss:ema_accuracy}

Table~\ref{tab:ema_hit_rate} reports the overlap between EMA's predictions and the true top-$K$ set on GLM-4-9B. The predictor achieves 98.05\% under Quest and 97.52\% under InfLLM-v2 across benchmarks.
A consistent trend is observed for other LLMs (\ref{ss:appdix_ema_hit_rate}).


\begin{table}[]
\centering
\footnotesize
\setlength{\tabcolsep}{1.3pt} 
\caption{The overlap rate(\%) between EMA's prediction and the true top-$K$ set of GLM-4-9B.}
\label{tab:ema_hit_rate}
\begin{tabular}{lcccccc}
\toprule
\textbf{Model} & \textbf{\shortstack{Long\\Bench}} & \textbf{\shortstack{Infinite\\Bench}} & \textbf{RULER} & \textbf{AIME} & \textbf{\shortstack{MATH\\500}} & \cellcolor{gray!20}\textbf{Avg.} \\
\midrule
Quest     & 98.65 & 96.86 & 98.36 & 98.15 & 98.25 & \cellcolor{gray!20}\textbf{98.05} \\ \hline
InfLLM-v2 & 97.86 & 97.14 & 96.46 & 98.15 & 98.01 & \cellcolor{gray!20}\textbf{97.52} \\
\hline
\end{tabular}
\end{table}

%% file: tex/related_work.tex
\section{Related Work}
\label{s:related_work}

\noindent\textbf{KV Cache Retrieval for Dynamic Sparse Attention.}
To avoid the information loss of KV-dropping methods~\cite{h2o-kvdropping,less-kvdropping,streamingllm-kvdropping}, KV retrieval approaches~\cite{infinigen,wu2025louiskv} keep the full KV cache in CPU memory and fetch only query-relevant tokens to reduce transmission, which is orthogonal to \systemname. LouisKV~\cite{wu2025louiskv} and AsyncSpade~\cite{luo2025asyncspade} predict the query state to prefetch relevant tokens, but approximate query-state estimation can drop important tokens and degrade quality. \systemname instead prefetches based on the temporal locality of historical importance scores and speculatively computes attention, with missed tokens reincorporated via attention repair for zero accuracy degradation.

\noindent\textbf{Temporal Locality for KV Cache Management.}
FlexiCache~\cite{takbir2026flexicache} exploits temporal stability \emph{across heads}: stable heads retain only their top-$K$ KV pages on GPU, while unstable heads keep all pages. This is complementary to \systemname, which exploits temporal locality \emph{across decoding steps} --- the tendency for the same blocks to be reselected at consecutive steps. The two are composable: \systemname's EMA predictor can guide which pages to retain within FlexiCache's stable heads, while \systemname's speculative attention separately improves decoding throughput.

\section{Conclusion}
We have introduced \systemname, a novel speculative attention runtime that breaks the selection-to-attention dependency in DSA.
To achieve this, we propose a lightweight EMA-based predictor and implement a customized sparse attention CUDA kernel that enables incremental attention repair.
Experiments on diverse LLMs and benchmarks show that \systemname achieves consistent speedups over serial DSAs.